\documentclass[3p,12pt]{elsarticle}

\usepackage{amssymb}

\usepackage{hhline}
\usepackage{threeparttable}
\usepackage{multirow}
\usepackage{booktabs}
\usepackage{multicol}
\usepackage[ruled,vlined]{algorithm2e}
\usepackage[figuresright]{rotating} 
\usepackage{graphicx}
\usepackage{colortbl}
\usepackage{color}
\usepackage{lineno}
\usepackage{hyperref}
\usepackage{float}
\usepackage{subfig}
\usepackage{tabularx}
\usepackage{setspace}

\journal{Review}

\begin{document}

\begin{frontmatter}

\title{Ensemble Reinforcement Learning: A Survey}

\author[]{Yanjie Song\fnref{fn1}}
\ead{songyj\_2017@163.com}
\author[]{Ponnuthurai Nagaratnam Suganthan\fnref{fn2} \corref{cor1}
}
\ead{p.n.suganthan@qu.edu.qa}
\author[]{Witold Pedrycz\fnref{fn3,fn4,fn5} \corref{cor2}
}
\ead{wpedrycz@ualberta.ca}
\author[]{Junwei Ou\fnref{fn1}}
\ead{junweiou@163.com}
\author[]{Yongming He\fnref{fn1}}
\ead{heyongming10@hotmail.com}
\author[]{Yingwu Chen\fnref{fn1}}
\ead{ywchen@nudt.edu.cn}
\author[]{Yutong Wu\fnref{fn6}}
\ead{wuyutong119@gmail.com}

\cortext[cor1]{Corresponding author}
\cortext[cor2]{Corresponding author}
\fntext[fn1]{College of Systems Engineering, National University of Defense Technology, Changsha, China}
\fntext[fn2]{KINDI Center for Computing Research, College of Engineering, Qatar University, Doha, Qatar}
\fntext[fn3]{Department of Electrical \& Computer Engineering, University of Alberta, Edmonton AB, Canada}
\fntext[fn4]{Systems Research Institute, Polish Academy of Sciences, Poland}
\fntext[fn5]{Faculty of Engineering and Natural Sciences, Department of Computer Engineering, Turkiye}
\fntext[fn6]{Department of Analytics, Operations and Systems, University of Kent, UK}

\begin{abstract}
Reinforcement Learning (RL) has emerged as a highly effective technique for addressing various scientific and applied problems. Despite its success, certain complex tasks remain challenging to be addressed solely with a single model and algorithm. In response, ensemble reinforcement learning (ERL), a promising approach that combines the benefits of both RL and ensemble learning (EL), has gained widespread popularity. ERL leverages multiple models or training algorithms to comprehensively explore the problem space and possesses strong generalization capabilities. In this study, we present a comprehensive survey on ERL to provide readers with an overview of recent advances and challenges in the field. Firstly, we provide an introduction to the background and motivation for ERL. Secondly, we conduct a detailed analysis of strategies such as model selection and combination that have been successfully implemented in ERL. Subsequently, we explore the application of ERL, summarize the datasets, and analyze the algorithms employed. Finally, we outline several open questions and discuss future research directions of ERL. By offering guidance for future scientific research and engineering applications, this survey significantly contributes to the advancement of ERL.
\end{abstract}

\end{frontmatter}

\section{Introduction}


Over the past several decades, reinforcement learning (RL) methods have proven to be highly effective in solving complex problems across various fields, including gaming, robotics, and computer vision. With the emergence of breakthroughs such as deep Q neural networks \cite{mnih2015human}, AlphaGo \cite{silver2016mastering}, video games \cite{vinyals2019grandmaster, kaiser2020model}, and robotic control tasks \cite{liu2021deep}, RL has witnessed a revitalization that outperforms human performance. The success of this approach is attributed to the agent's ability to automate feature acquisition and accomplish end-to-end learning. Artificial neural networks (ANN) and gradient descent further enhance RL's exploration and exploitation capabilities, rendering it suitable for handling laborious manual work or challenging tasks.

\begin{figure}[htp]
\centering
\includegraphics[width=0.52\textwidth]{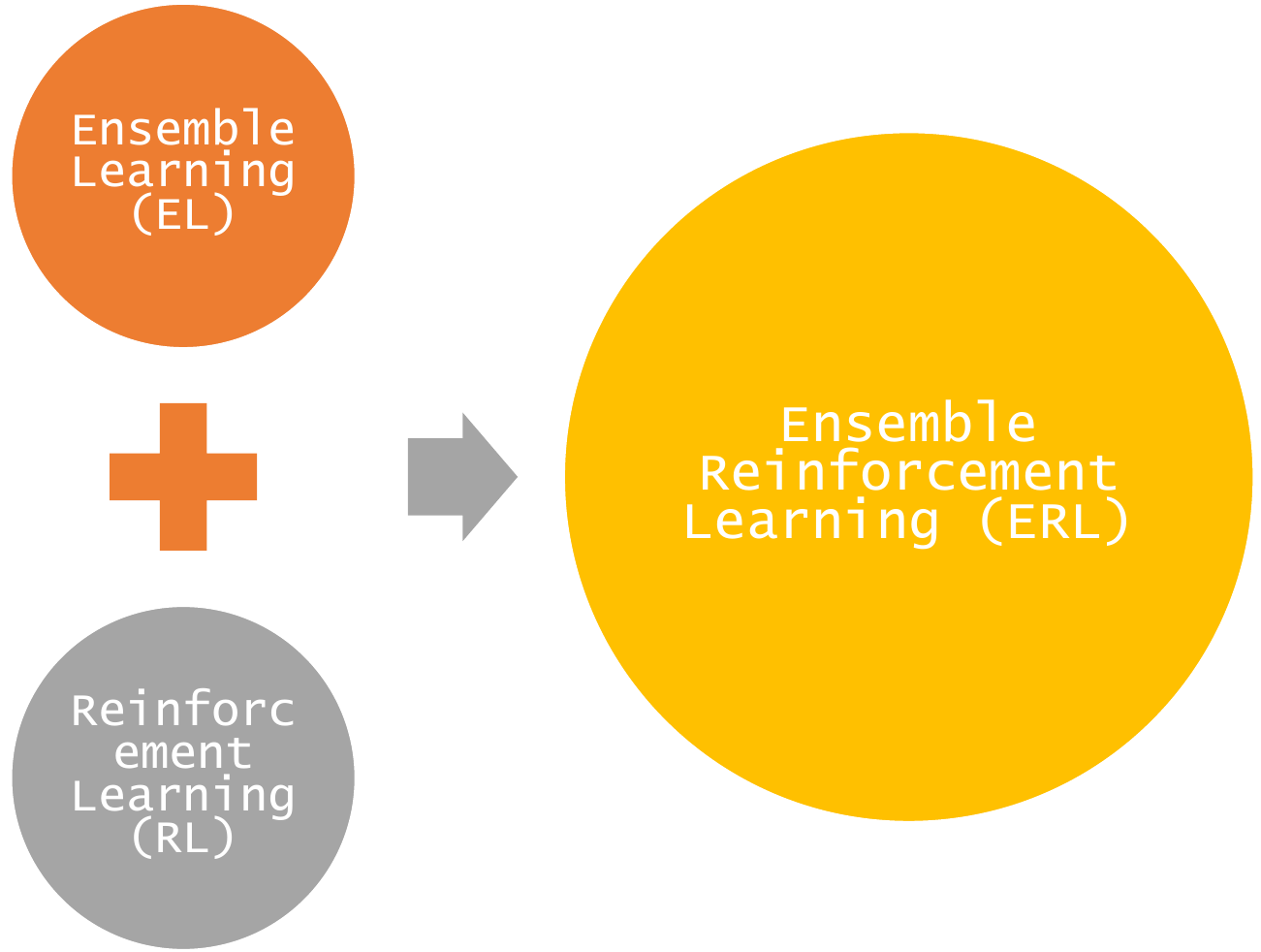}
\caption{\textcolor[rgb]{0,0,0}{Components of the ERL method. The two components EL and RL are combined to form ERL.}}
\label{Components of the ERL method}
\end{figure}

Nevertheless, each type of RL possesses distinct advantages and limitations. For instance, deep reinforcement learning (DRL) requires extensive training to obtain a policy \cite{kaiser2020model}, thereby introducing additional challenges such as overfitting \cite{fujimoto2018addressing}, error propagation \cite{kumar2019stabilizing}, and imbalance between exploration and exploitation \cite{chen2017ucb}. These challenges motivate researchers to design models or training algorithms. One approach is implementing ensemble learning (EL) into the RL framework, which enhances algorithmic learning and representation abilities (see Figure \ref{Components of the ERL method}). The method known as ensemble reinforcement learning (ERL) has demonstrated exceptional performance across various applications. The concept of EL was initially exemplified by Marquis de Condorcet \cite{condorcet1785essay}, who showed that average voting outperforms individual model decisions. The subsequent studies conducted by Krogh and Vedelsby \cite{krogh1995neural}, Breiman \cite{breiman2001random}, and other researchers have theoretically demonstrated the significant advantages of ensemble methods from various perspectives. The success of ensemble methods in the field of deep learning (DL) and RL can be attributed to three factors: the decomposition of datasets \cite{brown2005diversity}, powerful learning capabilities \cite{dietterich2000ensemble}, and diverse ensemble methods \cite{breiman2001random}.

\begin{figure}[htp]
\centering
\includegraphics[width=0.8\textwidth]{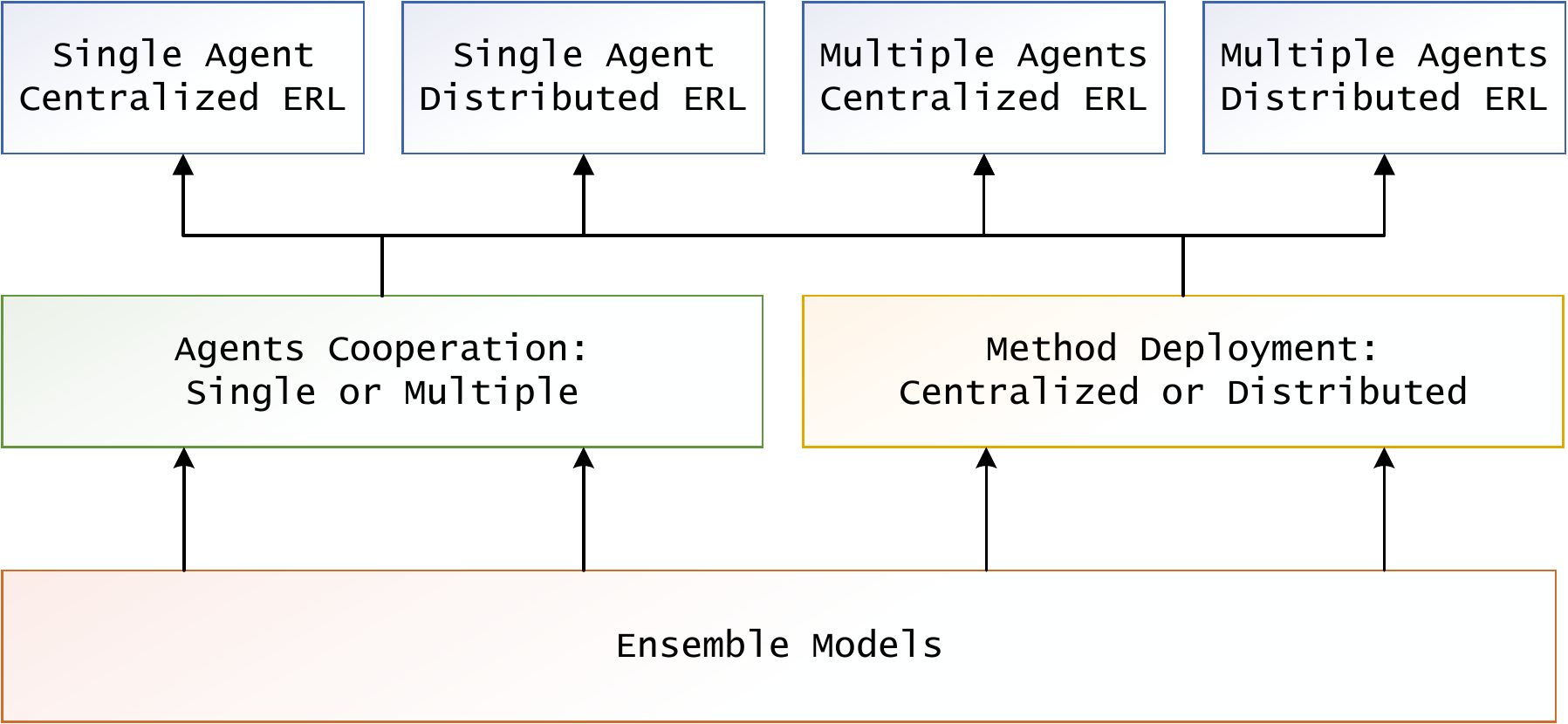}
\caption{A taxonomy of ERL according to agent cooperation and method deployment}
\label{A taxonomy of ERL according to agent cooperation and method deployment}
\end{figure}

The ERL method can be categorized according to different criteria. The constituent elements allow for the classification of ERL into high-level ensembles \cite{yang2020deep} and low-level ensembles \cite{sheikh2022maximizing}. ERL can also be classified as single-agent ERL \cite{chen2021randomized} and multi-agent ERL \cite{fausser2011ensemble} based on the number of agents involved. Moreover, centralized ERL \cite{anschel2017averaged} and distributed ERL \cite{jiang2021distributed} are classifications of ERL based on how the agents work. Figure \ref{A taxonomy of ERL according to agent cooperation and method deployment} presents a taxonomy according to agent cooperation and method deployment. All of these taxonomies are reasonable and can serve as reference frameworks for designing new ERL methods. The utilization of existing frameworks enables researchers to rapidly develop novel ERL methods. Additionally, comprehending the impact of strategies can assist researchers in directing their focus toward specific strategy design. In this paper, we provide a detailed description of ERL methods according to the improvement strategies used and discuss their applications to guide the design of new methods.

The literature on ERL encompasses a broad spectrum of related work, including training algorithms, ensemble strategies, and application domains. This paper aims to provide readers with a systematic overview of the existing research, current progress, and valuable conclusions achieved in this field. \textbf{To the best of our knowledge, this is the first survey focusing solely on ensemble reinforcement learning.} In this survey, we present the strategies employed in ERL and its associated applications, deliberate upon various unresolved inquiries, and offer a roadmap for future exploration in the realm of ERL. Approaching from this perspective enables readers to swiftly comprehend the ERL methodology while also facilitating the design and enhancement of tailored ERL approaches for specific problems or application scenarios.

\begin{figure}[htp]
\centering
\includegraphics[width=0.75\textwidth]{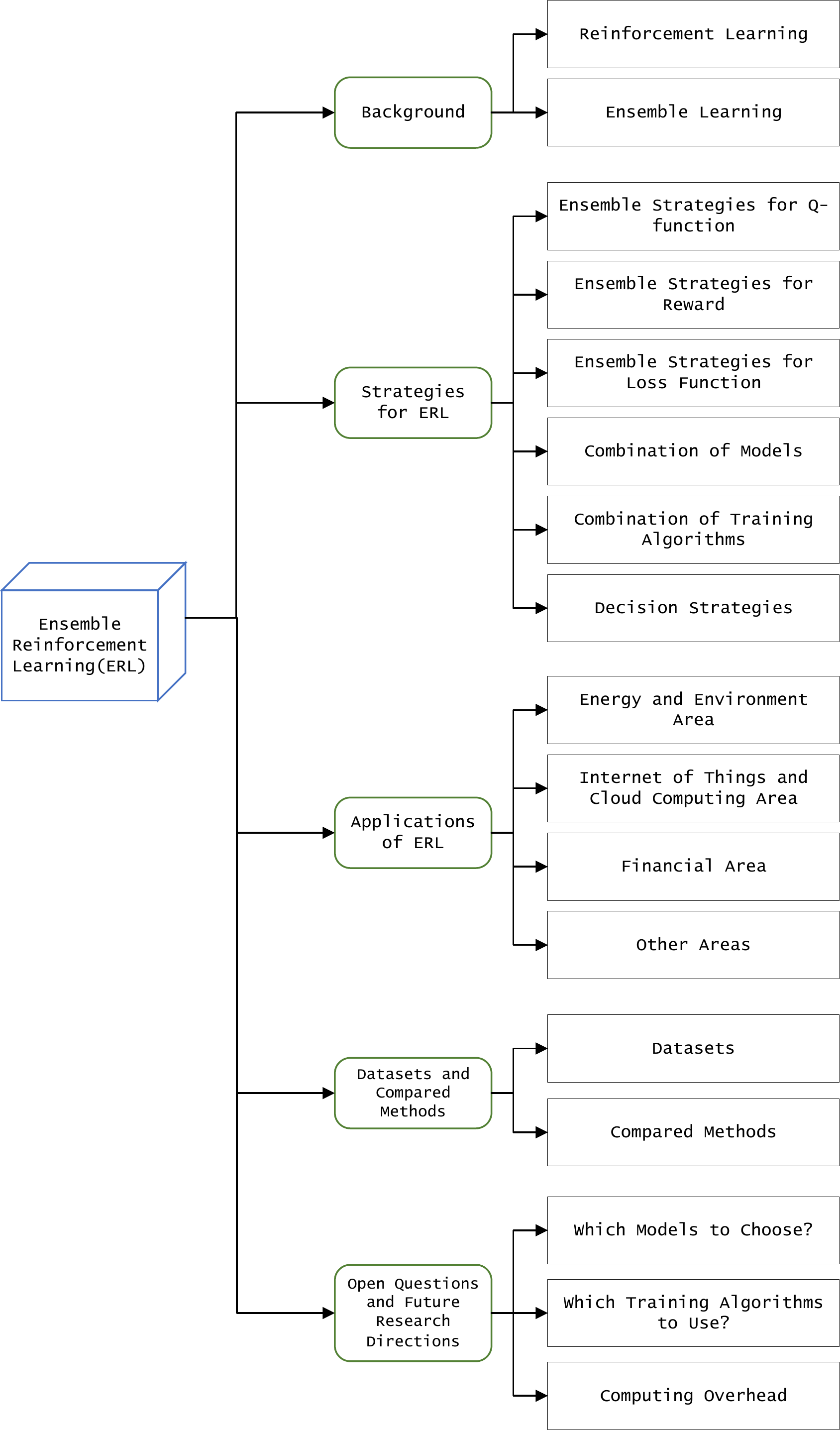}
\caption{Structure of the survey}
\label{Structure of the paper}
\end{figure}

The remainder of this paper is structured as follows. Section \ref{background} presents the background of ensemble reinforcement learning methods. Section \ref{strategy} introduces implementation strategies in ERL. Section \ref{application} discusses the application of ERL to different domains. Section \ref{application} discusses the datasets and compares methods used in the ERL-related studies. Section \ref{dataset} discusses several open questions and possible future research directions. Section \ref{conclusion} gives the conclusion of this paper. (See Figure \ref{Structure of the paper}).

\section{Background}
\label{background}

To enhance readers' comprehension of ensemble reinforcement learning methods, this section presents a concise overview of RL, EL, and ERL.

\subsection{Reinforcement Learning}

\begin{figure}[htp]
	\centering
	\includegraphics[width=0.5\textwidth]{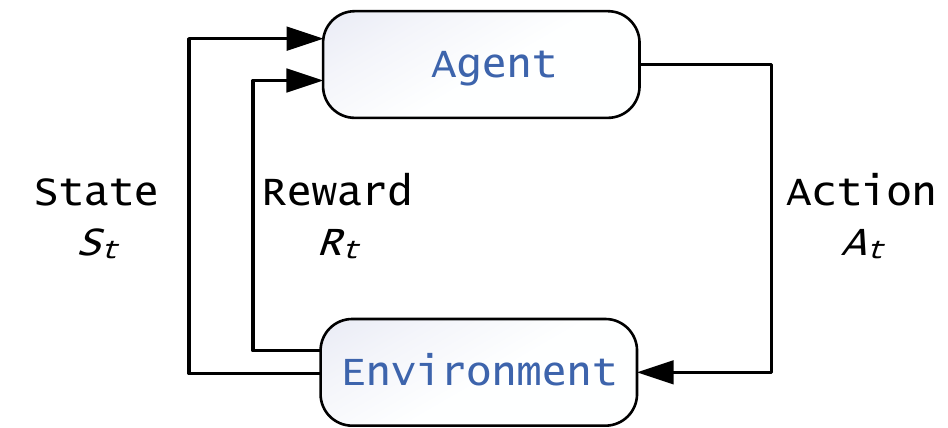}
	\caption{\textcolor[rgb]{0,0,0}{Interaction process between agent and environment. The Agent's performance is updated on the state through environmental evaluation.}}
	\label{Interaction process between agent and environment}
\end{figure}

Reinforcement learning is an artificial intelligence method in which an agent interacts with an environment and makes decisions iteratively to rectify errors, aiming to achieve optimal decision-making. \textcolor[rgb]{0,0,0}{Agent, the core of RL, is an entity that is capable of sensing the environment, making decisions, and taking actions.} Besides, the Markov Decision Process (MDP) forms the foundation for using RL to solve problems \cite{sutton2018reinforcement}. The RL approach is applicable when an agent's decision-making process is only related to the current state and not to the previous state. Figure \ref{Interaction process between agent and environment} (This figure is a correction to the figure published in the corresponding location of the ASOC article.) illustrates the agent-environment interaction process. A tuple $\langle S,A,P,R,\gamma\rangle$ can represent the MDP, where $S$ denotes the state, $A$ denotes the action, $P:S\times A\rightarrow P(S)$ denotes the state transfer matrix with the probability value $p(s^\prime\mid s)=p(S_{t+1}=s^\prime\mid S_t=s)$, $R:S\times A\rightarrow\mathbb{R}$ denotes the reward function, and $\gamma\in[0,1]$ denotes the discount factor. The agent's state at time step $t$ is $s_t$, and it will take the action $a_t$. The policy $\pi$ is defined by the combination of all states and actions, while the Q-value evaluates the expected reward obtained by the agent following policy $\pi$.
\begin{equation}
	Q^{\pi}(s,a)=\mathbb{E}_{\pi}\left[\sum\limits_{t=0}^{\infty}\gamma^{t}R(s_t,a_t)|s_0=s,a_0=a\right]
\end{equation}

The objective of using RL methods is to find an optimal policy $\pi$ that maximizes $Q^{\pi}$. For finite-state MDPs, Q-learning is the most prevalent RL method \cite{watkins1992q}, which uses a Q-table to record the combinations of $\langle$state,action$\rangle$. Subsequently, several RL methods incorporating artificial neural networks have been proposed to cope with the infinite state space.

Training algorithms can be categorized into model-based RL and model-free RL according to whether the environment model in RL is pre-defined or acquired through learning. Furthermore, these training algorithms can also be classified according to state-based, policy-based, or state-policy combination approaches. A more comprehensive account of the research progress on RL can be found in \cite{arulkumaran2017deep}.

\textcolor[rgb]{0,0,0}{The RL methods differ distinctly from the other classical classes of ML methods, namely supervised learning (SL) and unsupervised learning (UL), in several aspects. UL involves training a model using labeled datasets to enable the algorithm to predict accurate output labels based on input data. SL is primarily employed for regression and classification tasks. UL utilizes unlabeled data for model training to discover patterns, structures, or relationships within such data. Dimensionality reduction and clustering are representative UL techniques. As depicted in Table \ref{Differences between ERL, SL, UL}, these three types of methods exhibit significant differences across all three dimensions: data type used, feedback mechanism for the result, and target.}

\begin{table}[htbp]
	\centering
	\caption{Differences between ERL, SL, UL}
	\label{Differences between ERL, SL, UL}
	\tiny
	\begin{tabular}{l|l|l|l}
		\toprule[1.5pt]
		Dimension	& SL & UL & RL \\
		\midrule[1pt]
		Data type used & labeled & unlabeled & unlabeled \\
		Feedback mechanism for results & direct feedback & no feedback & multi-step post-implementation feedback \\
		Target & reduce error & find the hidden relationship & search the strategy with long-term reaward\\
		\bottomrule[1.5pt]
	\end{tabular}
\end{table}

\subsection{Ensemble Learning}

\begin{figure}[htbp]
	\centering
	\subfloat[Bagging \cite{breiman1996bagging} ]{\includegraphics[width=.4\textwidth]{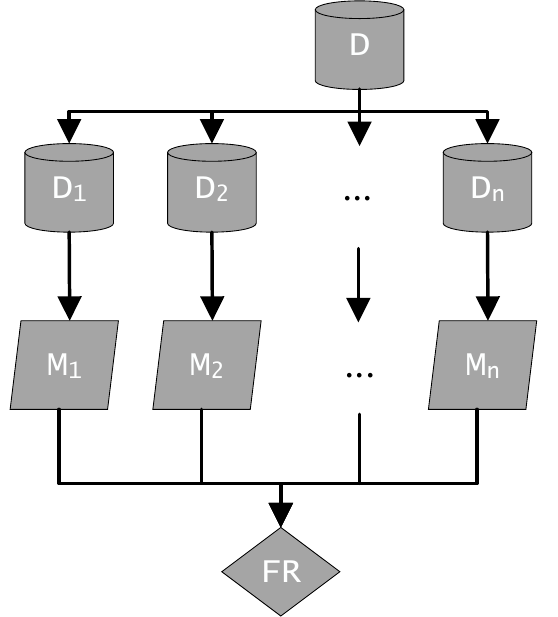}} \qquad
	\subfloat[Boosting \cite{schapire2003boosting} ]{\includegraphics[width=.4\textwidth]{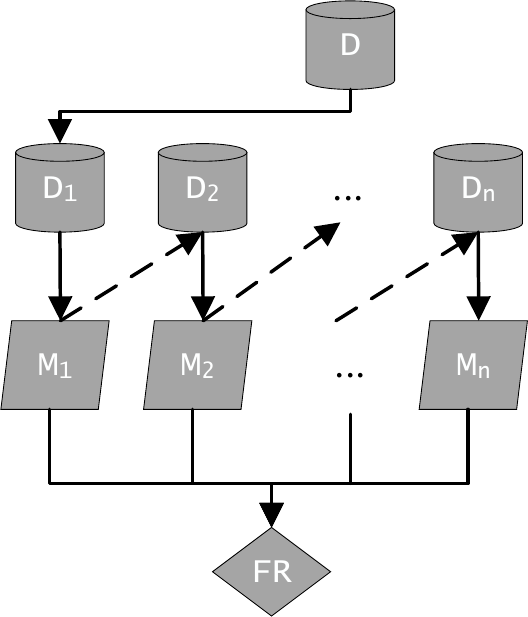}} \qquad
	\subfloat[Stacking \cite{wolpert1992stacked} ]{\includegraphics[width=.45\textwidth]{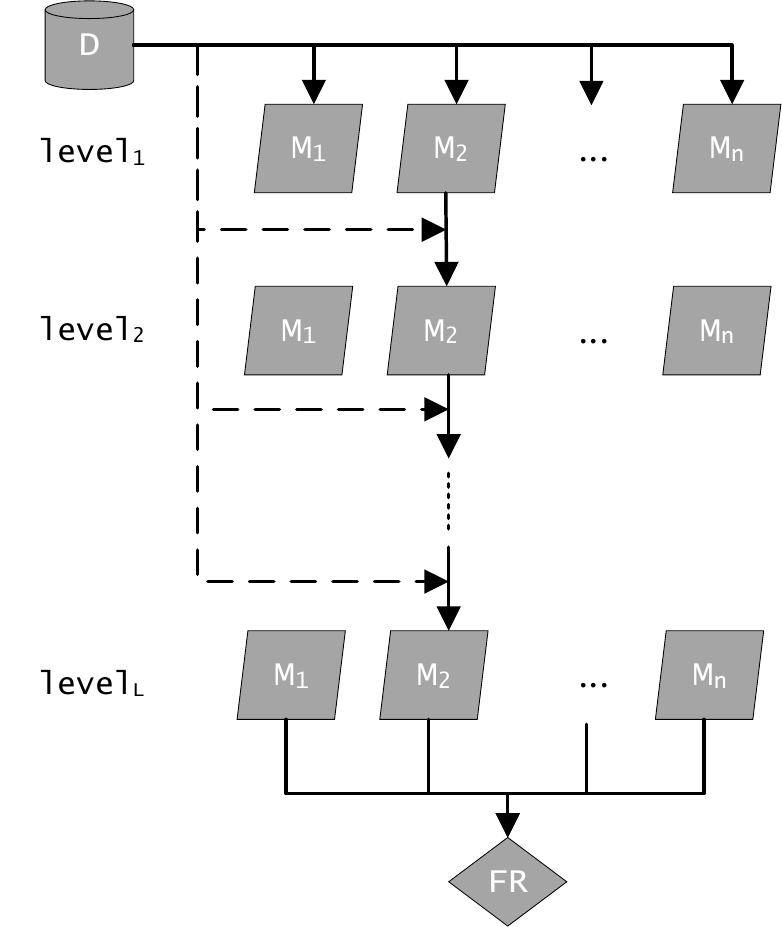}} \qquad
	
	\caption{\textcolor[rgb]{0,0,0}{Schematic diagram of three types of EL methods. (a) shows bagging. (b) shows boosting. (c) shows stacking.}}
	\label{Schematic diagram of three types of EL methods}
\end{figure}

Ensemble learning (EL) is a widely adopted approach in the field of machine learning (ML). The fundamental concept behind EL methods involves training multiple predictors, combining their outputs, and aggregating them to make informed decisions as the final result of an ensemble model. Compared to individual basic models, this EL method effectively leverages the distinctive characteristics of diverse model types to enhance the predictive performance and achieve more robust results. Prominent approaches in ensemble learning encompass bagging \cite{breiman1996bagging}, boosting \cite{schapire2003boosting}, and stacking \cite{wolpert1992stacked}. Figure \ref{Schematic diagram of three types of EL methods} gives a schematic diagram of these three types of EL methods, where $D$ denotes the dataset, $D_1$ to $D_n$ denote the sample selection from the dataset, $M_1$ to $M_n$ denote the models employed, and $FR$ denotes the final result. The dotted line in Figure \ref{Schematic diagram of three types of EL methods}-(b) indicates the dynamic nature of sample weights across subsequent iterations of the dataset. The dotted line in Figure \ref{Schematic diagram of three types of EL methods}-(c) indicates that all datasets are used for model prediction from level$_2$ to level$_L$. The primary distinction among these three types of methods lies in the approach to sample selection. These original and improved EL methods have been extensively employed across diverse domains, with the incorporation of domain knowledge in the improved EL method yielding exceptional performance outcomes. In summary, the EL method has demonstrated its advantageous nature through three key aspects.

$\bullet$ \textbf{Bias–variance Decomposition}

The bias-variance decomposition has been widely employed to demonstrate the effectiveness of ensemble learning (EL) methods over individual learning methods. While bagging reduces variance among base learners, other EL methods aim to reduce both bias and variance. Krogh and Vedelsby initially demonstrated the effectiveness of EL for single data set problems by employing ambiguity decomposition to reduce variance  \cite{krogh1995neural}. Subsequently, Brown et al. \cite{brown2005managing} and Geman et al. \cite{geman1992neural} verified the effectiveness of EL methods for multiple data set problems. The decomposition equation can be formulated as follows \cite{brown2005diversity}:
\begin{equation}\label{eq1}
	E[s-t]^2=bias^2+\frac{1}{N}var+(1-\frac{1}{N})covar
\end{equation}
\begin{equation}
	b i a s=\frac{1}{N}\sum_{i}(E[s_{i}]-t)
\end{equation}
\begin{equation}
	var=\frac{1}{N}\sum_i E[s_i-E[s_i]]^2
\end{equation}
\begin{equation}
	covar=\frac{1}{N(N-1)}\sum_i\sum_{j\neq i}E[s_i-E[s_i]][s_j-E[s_j]]
\end{equation}
where $ i $ denotes the $ i $-th model of EL, $ s $ denotes a solution to the problem, and $ N $ denotes the number of models in EL. The bias and variance are obtained using the average differences among multiple models, while $ covar $ measures the pairwise difference between models in the EL method.

The reduction in bias for an individual model is accompanied by an increase in variance. However, the ensemble model can be used for prediction purposes and effectively mitigate variance without increasing bias.

$\bullet$ \textbf{Statistical Perspective}

The advantages of EL from a statistical perspective are supported by the work conducted by Dietterich \cite{dietterich2000ensemble}. From a statistical point of view, machine learning problems exist within a search space encompassing multiple hypotheses. The target of the prediction model is to identify the optimal hypothesis. However, due to limited training data size relative to the expansive search space, there is an elevated risk of erroneous inferences. The use of an EL method can effectively integrate these hypotheses to enhance comprehension of the search space characteristics and mitigate the likelihood of erroneous classification or invalid prediction.

$\bullet$ \textbf{Diversity Perspective}

The advantages of EL from the diversity perspective are readily comprehensible and easily graspable. Dietterich highlights that the combination of different single models can enhance diversity \cite{dietterich2000ensemble}. Some typical EL methods, such as AdaBoost and random forest, show the importance of diversity in terms of training data. And the use of random noise can enhance the richness of the output. In other words, diversity allows decision-makers to combine the model output with usage requirements to obtain a more reasonable final result.

\textcolor[rgb]{0,0,0}{The integration of DL with EL, known as ensemble deep learning (EDL), has gained significant popularity in recent years. EDL demonstrates strong predictive capabilities by employing a model training approach that combines deep artificial neural networks (ANNs) and gradient descent \cite{nalepa2021deep,ganaie2022ensemble}. A comprehensive overview of the latest advancements in EDL can be found in \cite{ganaie2022ensemble}. It is noteworthy that ERL assesses the efficacy of model training based on environmental factors, whereas EDL relies on real-world datasets, emphasizing the distinction between ERL and EDL.}

\subsection{Ensemble Reinforcement Learning}

\begin{figure}[htp]
	\centering
	\includegraphics[width=0.75\textwidth]{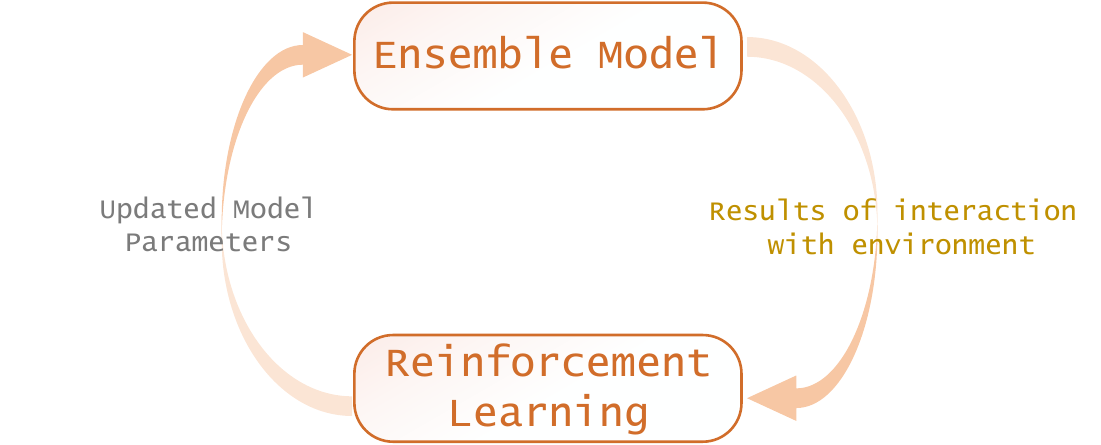}
	\caption{\textcolor[rgb]{0,0,0}{Data flow in ERL method. The information interaction will be continuous between RL and EL.}}
	\label{Data flow in ERL method}
\end{figure}

Ensemble reinforcement learning is a new artificial intelligence method that integrates reinforcement learning training methods into the ensemble learning framework, replacing conventional model training methods with more sophisticated RL methods. The training process of ERL involves a bidirectional exchange of data between the model and the training method. Figure \ref{Data flow in ERL method} illustrates the data flow between the ensemble model and reinforcement learning. The base learners in the ensemble model generate predictive data based on the task inputs, which interact with the environment and serve as the data source for the RL training. Once RL meets the training criteria, it consumes the data, leading to the generation of a new set of model parameters that are subsequently used to update the model. With the continuous generation and consumption of data, the ensemble model can effectively identify an optimal combination of parameters that can be suitably adapted to the environment. Then, the ERL method can directly leverage the trained ensemble model to solve complex tasks.

The structure of the ERL method exhibits a higher level of complexity compared to that of single EL or RL methods, thereby providing greater potential for enhancing method performance from various perspectives. In the next section, we will describe the improvement strategies in the ERL method in detail.

\section{Strategies for Ensemble Reinforcement Learning}
\label{strategy}
Previous studies have shown that the ERL method demonstrates superior average performance and sampling efficiency compared to RL methods, as evidenced by results obtained from public RL test sets and practical tasks \cite{sheikh2022maximizing, smit2021pebl}. By using ERL, the performance improvement can reach up to 20\% \cite{chengqing2023multi, shang2022new, tan2022new}. Moreover, for classification tasks, the ERL method achieves the best accuracy scores across multiple benchmarks in the UCI online data repository \cite{partalas2006ensemble,liu2020instance}.

The strategies employed by ERL to outperform other solution methods in various problems are closely interconnected. These strategies are categorized based on the diverse improvements made to the ERL. The ensemble strategies for ERL encompass Q-function, reward, and loss function ensembles, as well as model combinations, combination training algorithms, and decision strategies. In this section, we will introduce these strategies individually.

\subsection{Ensemble Strategies for Q-functions}

In most RL methods, the Q-function reflects how good the agent is in any given state \cite{sutton2018reinforcement}. A "good state" refers to a state that can achieve a high expected return, which is contingent upon the action taken by the agent. The background section provides the Q-function formula, which is applicable to ERL methods. Additionally, tailoring Q-functions specifically for ERL methods can further enhance the algorithm performance \cite{sheikh2022maximizing,hans2010ensembles,he2021mepg, an2021uncertainty}.

A Max-min Q-learning algorithm using multiple Q-functions was proposed by Lan et al. \cite{lan2020maxmin} to evaluate the performance. The max-min mechanism integrates multiple predicted values as a reference for agent decision-making. Specifically, the prediction term in the original Q-value calculation formula is determined by selecting the smallest value among multiple Q functions. The efficacy of this algorithm is demonstrated using the Mountain Car environment. It can be observed from the results that enhancing the Q-function positively impacts both the convergence performance and search efficiency of the algorithm.

The Q-function can also be enhanced through the utilization of Bayesian optimization. Chen et al. incorporated this concept into the design of the ERL method to update $ Q^* $ using Bayesian optimization, which proves particularly effective in addressing high-dimensional ERL problems and especially valuable when dealing with ultra-large solution spaces \cite{chen2017ucb}. In this study, an upper-confidence bounds (UCB) based strategy for exploring the solution space is employed for action selection by the agent. The proposed method's performance is evaluated using Atari games in the experimental section, and the results validate its effectiveness.

The ERL methods can leverage certain Q-value approximation techniques employed in RL research. Ghosh et al. proposed an ERL approach based on a multi-agent framework to address the air traffic control problem \cite{ghosh2021deep}. To expedite convergence, they utilized a kernel-based Q-value approximation method that utilizes sample transitions \cite{ormoneit2002kernel}.

\textcolor[rgb]{0,0,0}{The main targets of Q-function improvements can be summarized as follows: maintaining diversity \cite{sheikh2022maximizing,he2021mepg}, enhancing algorithmic exploration performance \cite{chen2017ucb}, and reducing bias (e.g., underestimation bias) or coping with the effects of overestimation \cite{lan2020maxmin}. This improvement strategy can be regarded as a key optimization after the integration of multiple single models, which allows the components to be integrated as a whole. This improvement strategy is more thorough and easier to get high-quality results than some other improvement strategies.}

\subsection{Ensemble Strategies for Reward}

The reward is a reflection of the agent's performance in taking actions based on the state. Generally, a high reward corresponds to a good decision, while a problematic decision prompts the agent to identify and rectify errors through the reward mechanism. Building upon this concept, Yao et al. proposed an averaging reward calculation method for the ERL method, enabling it to effectively balance exploration and exploitation \cite{yao2021sample}. Subsequently, an ANN model is trained using the soft actor-critic method. This ERL approach proves highly suitable for addressing challenges associated with exploring uncharted regions.

The combination of reward functions in ERL can also be utilized with weight aggregation. Lin et al. proposed an adaptive adjustment method for the weights of reward functions by combining Upper Confidence Bounds (UCB) and error \cite{lin2020ensemble}. This weight update strategy enables the ERL method to assess the accuracy of previous policies and enhance generalizability. Qi et al. also employed an ERL method with aggregated weighted reward functions to address the traffic signal control problem \cite{qi2022random}.

Although the traditional calculation method of reward is widely used, it has the shortcomings of a complex process. Compared to traditional methods, fuzzy-based methods can reduce computational costs. A fuzzy set can affect the reward value obtained by agents by measuring dissimilarity. Pan et al. proposed a dissimilarity evaluation metric for deciding the weight value of each agent's reward in ERL \cite{pan2019ensemble}. In this way, ERL can achieve a good training effect with fewer iterations.

\textcolor[rgb]{0,0,0}{Strategies for improving reward can modify the agent's action evaluation mechanism, thereby impacting both the state and adopted strategies. While most existing research focuses on processing feedback base learners in the environment and aggregating it to reward, limited studies specifically address comprehensive evaluation of differentiated performance among various base learners in ERL \cite{yao2021sample,pan2019ensemble}. Furthermore, designing novel methods for calculating rewards in ERL is also an improvement idea.}

\subsection{Ensemble Strategies for Loss Function}

The loss function serves as a fundamental foundation for ERL to update network parameters and enhance the performance of agent decisions. A smaller loss value indicates a closer proximity between the predicted value of the ERL model and the actual value. However, during RL model training, two critical issues often arise: gradient explosion and gradient disappearance. Several studies in ERL have endeavored to refine the accuracy of RL model decisions by enhancing the loss function \cite{sheikh2022maximizing,lee2022offline,yang2022towards,goyal2019reinforcement}. Kumar et al. conducted theoretical analysis on bootstrapping error and proposed an approach to reduce error accumulation, thereby augmenting stability in ensemble Q-learning algorithms \cite{kumar2019stabilizing}.

Designing a global loss function for all models used is another approach specific to ERL. Adebola et al. proposed an improved global loss function with each member model included in the function \cite{adebola2022deft}. Moreover, an interpolation method is used to control the difference between policies that the algorithm needs to train. This ERL method can also optimize the agent's policy selection using fine-tuning techniques. Based on this idea, Jiang et al. added the training data error between models to the overall loss calculation formula for improving prediction accuracy \cite{jiang2021distributed}. It can be seen from the experiment that this method is applicable in mobile edge computing (MEC) systems for rational resource scheduling. 

In addition, incorporating uncertainty into the analysis is an effective approach to enhance the loss function. Sun et al. employed the technique of uncertainty reduction to devise an ensemble loss function \cite{sun2020ensemble}, which effectively mitigates the risk of the RL model getting trapped in a local optimum. Within this study, a distillation method is utilized for selecting training for the model. Subsequently, the performance of the proposed ERL method is validated through Atari game experiments.

\textcolor[rgb]{0,0,0}{The improvement of the loss function can make the model prediction of ERL closer to the real situation \cite{sutton2018reinforcement}. However, due to the existence of bias and variance, it is not guaranteed that a high-precision model must guarantee excellent performance in processing tasks. This is particularly true for complex sequential decision problems where environmental changes significantly impact the agent's performance. Therefore, designing a robust loss function that ensures stable performance across various scenarios becomes crucial when considering further advancements in improving the ERL method.}

\subsection{Combination of Models}

The ensemble of different types of models is a common and simple strategy in ERL. \textcolor[rgb]{0,0,0}{The combination of models can achieve structural diversity.} These models can be either ML models or ANN models \cite{dong2021novel,perepu2020reinforcement}. The structure of the model combination is determined according to the specific problem and the solution target. A single type or a combination of multiple types of models are both popular. For using only one type of model, ANNs with different depths can be considered. While other studies use different random initialization strategies \cite{chen2021randomized} or ANN in different training stages \cite{carta2021multi}. Table \ref{Combination of models} provides a summary of related work ensemble model combination strategies. There are three models mainly in relevant works, including ML models only, ANN models only, and ML\&ANN models hybrid.

\textcolor[rgb]{0,0,0}{The model selection process typically involves choosing from a range of classical or recently proposed methods that are specifically designed for addressing this type of task. For instance, when it comes to prediction tasks, most researchers will choose convolutional neural networks, gated recursive units, artificial neural networks (ANN), and other SOTA approaches in this domain \cite{perepu2020reinforcement,liu2021new,elliott2021wisdom}. Apart from determining the potential composition of ERL, it is also crucial to determine the number of base learners. In most studies, two or three single models are commonly employed. Besides, there exist some studies where more than three models have been integrated within the ERL framework \cite{cao2022novel}. Saadallah et al. \cite{saadallah2021online}, Li et al. \cite{li2022deep}, and Sharma et al. \cite{sharma2022deepevap} are some examples in this regard. These studies have extensively demonstrated that employing a combination of strategies in ERL outperforms baseline methods as well as state-of-the-art ensemble learning techniques \cite{tan2022new,jalali2021newhybrid}.}

\begin{table}[htp]
\centering
\caption{Combination of models}
\label{Combination of models}
\small
\begin{tabularx}{\textwidth}{p{1cm}p{4cm}X}
\toprule[1.5pt]
Year &
Author &
Combination of Models \\
\midrule[1pt]
2019 &
Dong et al. \cite{dong2021novel} &
long short-term memory (LSTM) network,   gated recurrent unit network \\
2019 &
Goyal et al.   \cite{goyal2019reinforcement} &
convolution neural network (CNN),   gated recursive unit \\
2020 &
Liu et al.    \cite{liu2020new} &
LSTM network,   deep belief network, echo state network \\
2020 &
Perepu et al.  \cite{perepu2020reinforcement} &
linear regression model, LSTM model, ANN, random forest \\
2021 &
Liu et al.  \cite{liu2021new} &
graph convolutional   network, 
LSTM networks, gated recursive unit \\
2021 &
Saadallah et al.  \cite{saadallah2021online} &
autoregressive integrated moving   average, exponential smoothing, gradient boosting machines, gaussian  
processes, support vector regression, random forest, projection pursuit   regression, MARS, principal component regression, decision tree regression,  
partial least squares regression,   
multilayer perceptron, LSTM network, Bi-LSTM:  
Bidirectional LSTM, CNN-based LSTM, convolutional LSTM \\
2021 &
Daniel L. Elliott and Charles   Anderson \cite{elliott2021wisdom} &
CNN,   gated recursive unit, ANN \\
2022 &
Shang et al. \cite{shang2022new} &
gated recursive unit, graph  convolutional network, graph attention network \\
2022 &
Tan et al.    \cite{tan2022new} &
graph attention network, long  short-term memory networks, temporal convolutional network \\
2022 &
Li et al.  \cite{li2022new} &
gated recurrent unit,  deep belief network, temporal convolutional   network \\
2022 &
Zijie Cao and Hui Liu   \cite{cao2022novel} &
temporal convolutional network,   Bidirectional long short‑term memory network, kernel extreme learning machine \\
2022 &
Birman et al.   
\cite{birman2022cost} &
machine learning models,   ANN \\
2022 &
Li et al.  \cite{li2022deep} &
naive bayes, support vector   machine with stochastic gradient descent, FastText, Bi-directional LSTM \\
2022 &
Sharma et al.  \cite{sharma2022deepevap} &
support vector regressor (SVR),   eXtreme gradient boosting (XGBoost), Random Forest (RF), ANN, LSTM, CNN, CNN-LSTM, CNN-XGB, CNN-SVR, and CNN-RF \\
2022 &
Shi Yin and Hui Liu   \cite{yin2022wind} &
group method of data handling,   echo state network, extreme learning machine \\
2023 &
Yu et al.  \cite{chengqing2023multi} &
graph attention network, gated   recursive unit, temporal convolutional network\\
\bottomrule[1.5pt]
\end{tabularx}
\end{table}

\begin{figure}[htp]
\centering
\includegraphics[width=0.65\textwidth]{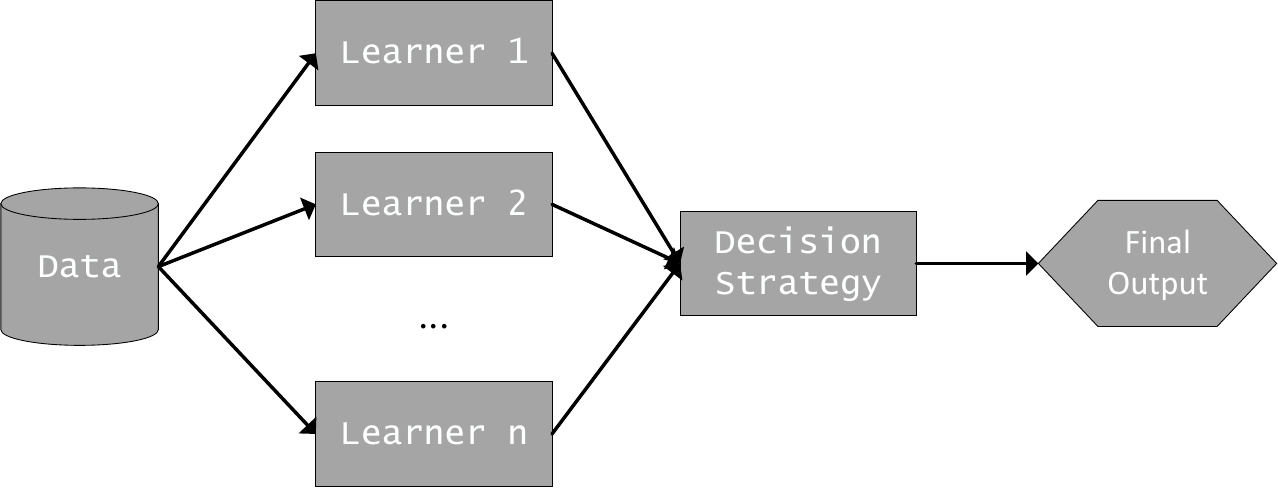}
\caption{\textcolor[rgb]{0,0,0}{Parallel ensemble reinforcement learning. The respective results of base learners are processed by the decision strategy to get the final output.}}
\label{Parallel ensemble reinforcement learning}
\end{figure}

\begin{figure}[htp]
\centering
\includegraphics[width=0.65\textwidth]{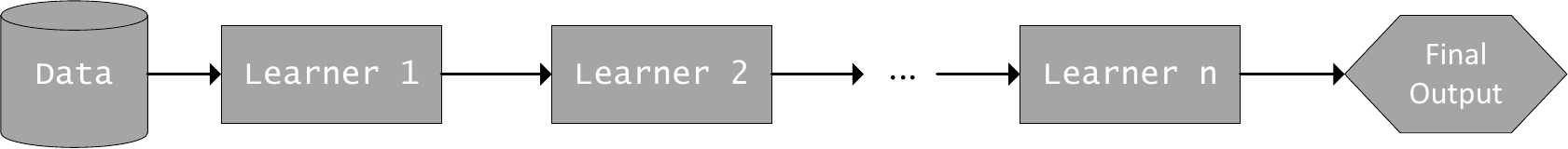}
\caption{\textcolor[rgb]{0,0,0}{Sequential ensemble reinforcement learning. The output of the previous base learner will be used as the input of the following base learner.}}
\label{Sequential ensemble reinforcement learning}
\end{figure}

ERL can be divided into parallel ERL and sequential ERL according to the relationship between base learners in ERL. Figure \ref{Parallel ensemble reinforcement learning} and Figure \ref{Sequential ensemble reinforcement learning} give schematic diagrams of these two ERL methods. In most ERL studies, such as Liu et al. \cite{liu2020instance}, Schubert et al. \cite{schubert2022polter}, and Shen et al. \cite{shen2019robust}, base learners are constructed in a parallel framework. These ML or ANN models are responsible for the same task. After each model processing, the final prediction result will be generated by a certain strategy. There are also some studies, such as Qin et al. and Ferreira et al., that try to construct the ERL method in the sequential framework \cite{qin2022hrl2e,ferreira2018multiobjective}. In this framework, the base learner completes the final prediction step by step in a certain order \cite{ferreira2018multiobjective}. 

\textcolor[rgb]{0,0,0}{Model combination is a readily implementable strategy for enhancing ERL performance. By integrating multiple classical or advanced models, the diversity of ERL methods can be maintained while leveraging the strengths of each model \cite{elliott2021wisdom,sharma2022deepevap}. However, it does not necessarily follow that increasing the number and variety of models in ERL will always lead to improved performance \cite{cully2017quality}. The single models and the design of decision mechanisms significantly impact final outputs. Moreover, due to the numerous parameters involved, training these models requires substantial data and time. Therefore, achieving a balance between the number of models employed, performance enhancement, and training costs becomes a key consideration when adopting the model combination strategy.}

\subsection{Combination of Training Algorithms}

\begin{table}[htp]
\centering
\caption{Combination of training algorithms}
\label{Combination of training algorithms}
\begin{tabularx}{\textwidth}{p{1cm}p{4cm}X}
\toprule[1.5pt]
Year &
Author &
Combination of Training   Algorithms \\
\midrule[1pt]
2008 &
Marco A. Wiering and Hado van   Hasselt \cite{wiering2008ensemble} &
Q-learning, Sarsa, actor-critic,   QV-learning, ACLA \\
2018 &
Chen et al.   \cite{chen2018ensemble} &
deep Q-networks, deep Sarsa networks, double deep Q-networks \\
2020 &
Yang 
et al. \cite{yang2020deep} &
proximal policy optimization,   advantage actor-critic, deep deterministic policy gradient \\
2020 &
Saphal et al.   \cite{saphal2020seerl} &
advantage actor-critic, sample   efficient actor-critic with experience replay, actor-critic using 
Kronecker-factored trust region, deep deterministic policy gradient, soft  
actor-critic, trust region policy optimization \\
2021 &
Smit et al.  \cite{smit2021pebl} &
double deep Q-Learning, soft   actor-critic \\
2022 &
Eriksson et al.  \cite{eriksson2022sentinel} &
residual gradient, TD, TD$ \left(\lambda\right) $ \\
2022 &
N{\'e}meth, Marcell and Sz{\H{u}}cs, G{\'a}bor \cite{nemeth2022split} &
deep deterministic policy   gradient, advantage actor-critic, proximal policy optimization\\
\bottomrule[1.5pt]
\end{tabularx}
\end{table}

Ensemble Reinforcement Learning (ERL) can not only use model combinations to obtain diverse prediction results but can also use different training algorithms to achieve full exploration of the solution space. \textcolor[rgb]{0,0,0}{Parameter diversity can be achieved by using multiple training algorithms to get the respective parameters of base learner.} Training algorithms can be classified into three categories: state-based, policy-based, and state-policy combination-based. Each of these training algorithms has its unique sampling strategy and output data characteristics. Researchers can quickly use the training algorithms according to the application scenarios without focusing on data sampling technology, which is similar to the EL method \cite{wiering2008ensemble,nemeth2022split}. Table \ref{Combination of training algorithms} provides information about studies using the combination strategy of training algorithms. There exists a new method of combining online and offline training algorithms or using training algorithms based on different optimization strategies, which can take advantage of their respective strengths to handle complex tasks. Accordingly, the complexity of ERL methods using such improved strategies increases. For this reason, the training process of the ERL method takes more time. Moreover, the ensemble model obtained also requires the design of a decision strategy to select the prediction results that are closest to the actual situation.

Currently, the research related to the combination of training algorithms is not deep enough and simply combines multiple typical algorithms. \textcolor[rgb]{0,0,0}{The combination of training algorithms in the ERL, similar to the model combination strategy, increases parameters (primarily hyperparameters). However, intricate combinations of training algorithms can result in significant time investment for implementation and debugging \cite{wang2021evolutionary}. Moreover, excessive hyperparameters may lead to models lacking robustness across different environments \cite{zhang2021importance}.} It is worth noting that there are some similarities between training algorithms. Transfer learning can be considered to transfer sampled data to reduce the training time. In addition, the termination conditions of multiple training algorithms also deserve in-depth analysis. Using the same number of iterations may result in some models finding the best policy long ago, while some models still need further training. Therefore, more research is required to investigate the combination of training algorithms in ERL.

\subsection{Decision Strategies}
\begin{figure}[htp]
\centering
\includegraphics[width=0.5\textwidth]{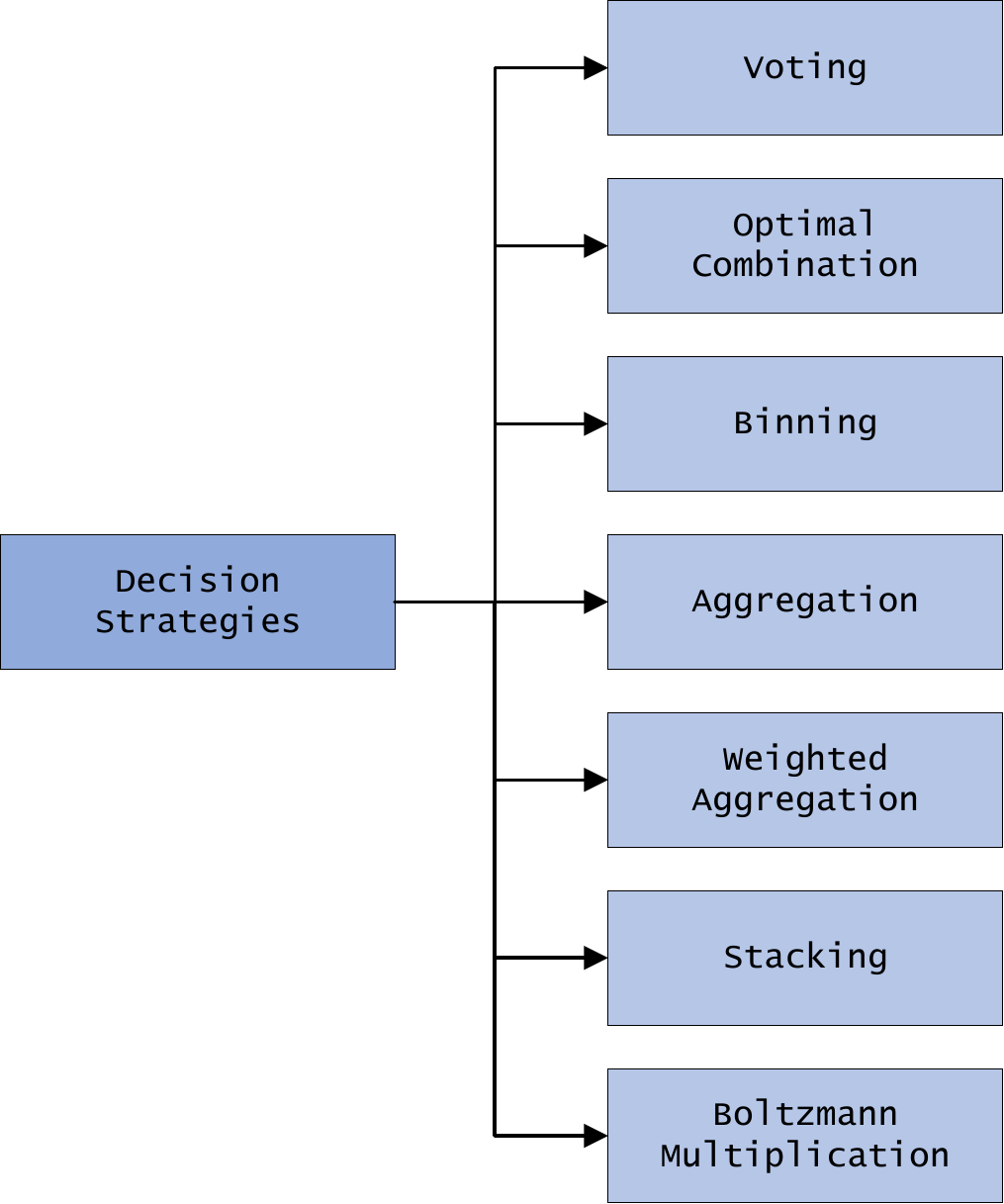}
\caption{\textcolor[rgb]{0,0,0}{Decision strategies. Decision strategies mainly include: voting, optimal combination, binning, aggregation, weighted aggregation, stacking, and Boltzmann multiplication.}}
\label{Decision strategies}
\end{figure}

The ERL algorithm employs multiple base learners to generate individual results which may introduce variations among the outputs. Consequently, specific decision strategies are necessary for the ERL model and output. Commonly adopted decision strategies in existing relevant research encompass voting, optimal combination, binning, aggregation, weighted aggregation, stacking, and Boltzmann multiplication (refer to Figure \ref{Decision strategies}).

\textbf{Voting}: Voting, as a common ERL decision strategy, records the number of occurrences of each prediction at first \cite{fausser2015neural}. Then, the final prediction can be selected from the results according to the principle of majority or ranking.

\textbf{Optimal Combination}: For classification problems, this is a commonly used decision strategy \cite{jalali2021oppositional}. Multiple base classifiers are trained separately, from which the optimal subset of models is selected to form an ensemble to classify the test set.

\textbf{Binning}: Binning is a majority voting decision strategy with continuous action space \cite{saphal2020seerl}. First, the action space is discretized into multiple intervals. Then, the number of occurrences of the actions in each interval is recorded. Finally, the average value of the action within the interval with the highest number of occurrences is selected as the final prediction result.

\textbf{Aggregation}: The prediction results of all the models in ERL are summed to produce an overall evaluation value, which is taken as the final result \cite{li2022deep}. In the aggregation method, each model is considered to be equally reliable.

\textbf{Weighted Aggregation}: The prediction results obtained from different models are summed according to their weights \cite{perepu2020reinforcement}. A high value of weight is used for aggregation models with high prediction accuracy.

\textbf{Stacking}: First, an additional machine learning model is involved to further predict the results of base learners. Then, the output of this machine learning model is used as the final prediction result \cite{birman2022cost}.

\textbf{Boltzmann Multiplication}: Boltzmann distribution is the basis for decision making \cite{wiering2008ensemble}. The probability value of each action can be calculated according to the Boltzmann distribution. The outcome with the highest probability value is selected and will not be changed.

In summary, the selection of decision strategies in ERL depends on the specific application scenario and the characteristics of the base learners. At present, there is no related study on the in-depth analysis of the application scenarios of these strategies. The study in this area will be helpful to improve the prediction accuracy of ERL methods. It will also promote the process of ERL research.

\subsection{Discussions}

\textcolor[rgb]{0,0,0}{In recent years, a variety of improvement strategies have emerged to enhance the performance of ERL. These include adjusting the Q-function, reward, and loss function in RL, designing model combinations, training algorithm combinations, and decision-making mechanisms. These strategies have improved ERL's ability to solve complex tasks from various aspects and promoted its application in different fields. However, the rapid development of this method has exposed a deficiency in insufficiently in-depth theoretical analysis of ERL improvement strategies. Most studies only focus on improving ERL from an application perspective without analyzing the motivation for adopting these strategies. This situation may lead to numerous similar works that do not advance the field well. Therefore, we aim to discuss two important issues here with hopes of drawing attention from other researchers.}

\textcolor[rgb]{0,0,0}{The first problem is how to select strategies to improve the ERL. Irrespective of the method employed for strategy selection, the objective remains consistent - improving ERL performance for task-solving purposes. These strategies vary significantly in terms of the effort required to adapt the approach due to varying levels of ERL modifications involved. For instance, proposing a new Q-function or loss function necessitates substantial knowledge and experience in RL and ERL methods. This strategy design approach demands considerable time and effort from researchers. However, it offers an opportunity to apply the newly proposed ERL across a range of problem scenarios or tasks, thereby driving research development forward. Many researchers facing challenges when directly designing new mechanisms for RL find it easier to combine or enhance existing techniques while applying them to novel problems. Although they may encounter difficulties when attempting to direct in other areas, this type of research effectively addresses their specific interests. There is an urgent need for systematic analysis of how different strategies impact ERL improvement and obtain applicable improvement strategies across various scenarios or problems. Such systematic analysis will foster rapid development and widespread implementation of ERL within diverse fields. Additionally, analyzing the effects of multiple strategy combinations is essential in order to fully realize the potential of ERL.}

\textcolor[rgb]{0,0,0}{Another question to consider is whether more complex structures are superior. Not only the ERL field but also other ML fields exhibit a trend towards increasingly intricate models and algorithms. While complex structures can indeed enhance method performance, they inevitably lead to greater computational resource consumption \cite{he2021mepg}. Additionally, the complexity of modeling structures may result in numerous hyperparameters requiring researchers to train effective models. Furthermore, these models may experience high volatility in performance when applied to different scenarios \cite{zhang2021importance}. Therefore, it is crucial to design a sound strategy that achieves desired goals by analyzing the impact of increasing model numbers and types integrated into ERL method performance enhancement. Researchers should focus more on ERL strategy design if an appropriate number of models are available.}

\section{Applications of Ensemble Reinforcement Learning}
\label{application}
\begin{figure}[htp]
\centering
\includegraphics[width=0.65\textwidth]{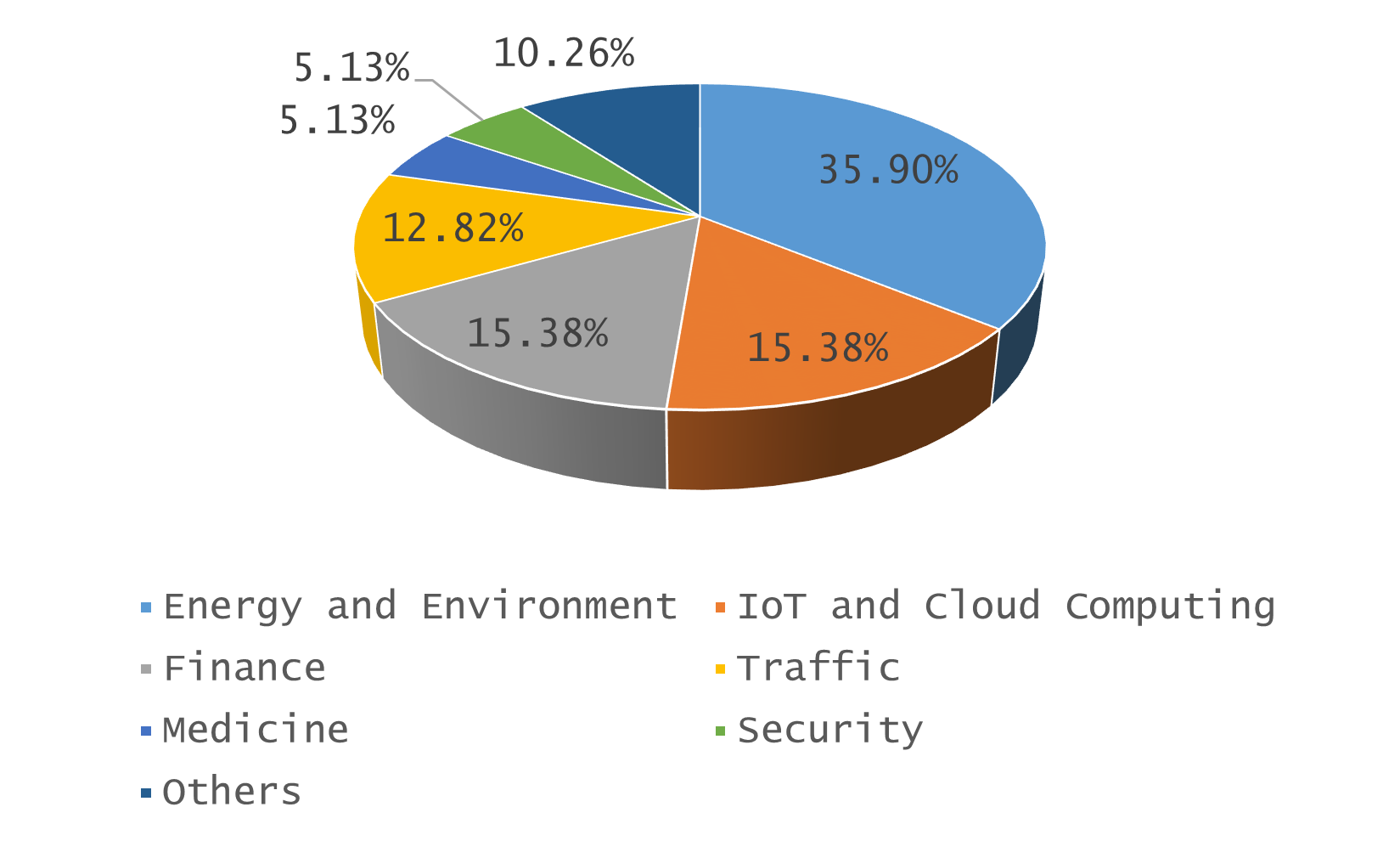}
\caption{\textcolor[rgb]{0,0,0}{The summary of the different application areas of ERL. The area where ERL is most applied is energy and environment.}}
\label{The summary of the different application areas of ERL}
\end{figure}

A significant portion of existing research on ERL primarily focuses on discrete/continuous control actions \cite{lin2020ensemble,wu2020deep,sheikh2022dns,buckman2018sample} and game environments \cite{chen2018effective,peer2021ensemble,brown2018interpretable} to verify the effectiveness of proposed algorithms. Additionally, researchers attempted to utilize ERL methods to solve in practical domains. Figure \ref{The summary of the different application areas of ERL} provides an overview of the key application areas encompassing energy and environment, IoT and cloud computing, finance, and other sectors where ERL has been extensively explored. Among these domains, energy and environment emerge as the most extensively investigated area for applying ERL techniques. In this section, we discuss the application of ERL in various domains.

\subsection{Energy and Environment Area}

As the global economy continues to expand, energy and environmental issues have gained increasing attention worldwide. The utilization of neural network methods for predicting future conditions based on historical data has become the policy formulation. In this type of prediction problem, there exists a spatiotemporal relationship between data, which makes the recurrent neural network (RNN) the preferred choice. In these related studies, ensemble RNN models (ERL) are employed to obtain. Table \ref{Application in energy and environment area} presents recent applications of ERL in the field of energy and environment. It is evident that wind power and PM 2.5 prediction are prominent research topics. From a statistical perspective, sixty-four percent of ERL-based studies utilized Q-learning as the training algorithm, while the remaining studies employed Sarsa, deep Q-network, and deep deterministic policy gradient algorithms. Overall, ERL methods in the energy and environment domain primarily focus on ensemble models where reinforcement learning algorithms are predominantly used directly. Compared with traditional approaches employing ML or ANN prediction methods \cite{pak2020deep,ma2020application}, there exists a significant gap between their respective prediction results and those obtained by ERL.

\begin{table}[htp]
\centering
\caption{Application in energy and environment area}
\label{Application in energy and environment area}
\begin{tabularx}{\textwidth}{p{1cm}p{4cm}XX}
\toprule[1.5pt]
Year & Authors                                               
& Problem                                   & Training Algorithm                 \\
\midrule[1pt]
2020 & Liu et al.  \cite{liu2020new}                  & wind speed short term   forecasting       & Q-learning                         \\
2021 & Jalali et al.  \cite{jalali2021new}            & solar irradiance   forecasting            & Q-learning                         \\
2021 & Liu et al.  \cite{liu2021new}                  & PM2.5 forecasting                        & Q-learning                         \\
2021 & Li et al.  \cite{li2021novel}                  & PM2.5 forecasting                         & Q-learning                         \\
2021 & Chao Chen and Hui Liu   \cite{chen2021dynamic} & wind speed prediction                    
& deep Q-Network                    
\\
2021 & Jalali et al.  \cite{jalali2021newhybrid}      & wind power forecasting                    & Q-learning                         \\
2022 & Tan et al.  \cite{tan2022new}                  & PM2.5 prediction                          & Sarsa                              \\
2022 & Qin et al.    \cite{qin2022optimization}       & unit commitment problem                   & deep Q-Network                     \\
2022 & Sogabe et al.  \cite{sogabe2022attention} & smart energy optimization and   risk evaluation & Q-learning \\
2022 & Sharma et al.  \cite{sharma2022deepevap}       & estimating reference   evapotranspiration & Q-learning                         \\
2022 & He et al.  \cite{he2022ensemble}               & wind farm control                         & deep deterministic policy gradient \\
2022 & Jalali et al.  \cite{jalali2022solar}          & solar irradiance   forecasting            & Q-learning                         \\
2022 & Shi Yin and Hui Liu   \cite{yin2022wind}       & wind power prediction                     & Q-learning                         \\
2023 & Yu et al.  \cite{chengqing2023multi}           & wind power prediction                     & deep deterministic policy gradient\\
\bottomrule[1.5pt]
\end{tabularx}
\end{table}

\textcolor[rgb]{0,0,0}{The most commonly employed improvement strategy in related studies, such as \cite{chengqing2023multi,jalali2021newhybrid,yin2022wind}, involves the combination of multiple models and decision strategy design. By combining models with different structures, ERL enables diverse decision-making and achieves structural diversity. For instance, \cite{tan2022new} integrates three distinct classes of spatio-temporal ANNs, graph attention network (GAT), LSTM networks, and temporal convolutional network (TCN), simultaneously for prediction. These ANNs employ different processing logics, effectively ensuring distinguishable output results.}

\textcolor[rgb]{0,0,0}{The application of ERL in the field of energy and environment primarily involves forecasting tasks, for which evaluation metrics used in traditional forecasting tasks are employed. To ensure an accurate assessment of the predictive performance of this method, mean absolute error (MAE), mean absolute percentage error (MAPE), and root-mean-square error (RMSE), are utilized \cite{liu2021new}. In addition, the standard deviation of error (SDE) is also a common metric for evaluating ERL methods \cite{li2021novel}.}

\subsection{Internet of Things and Cloud Computing Area}

In the area of the Internet of Things (IoT) and cloud computing, ERL is widely used to optimize system performance and business processing capabilities. The IoT connects various devices such as sensors, smart terminals, and industrial systems to form a globally interconnected system. Optimizing the efficiency of these devices and facilities has a positive impact on improving the overall performance of IoT systems. Cloud computing is another technology closely related to the IoT. Users can access computing resources or services in this distributed system provided by a cloud platform over the network on demand. Resource allocation and optimization have been the focus of research in the IoT and cloud computing area. Table \ref{Application in IoT and cloud computing area} presents the applications of ERL methods in this area. \textcolor[rgb]{0,0,0}{Among these related works, the diversity of ERL in \cite{gu2021heterogeneous} is primarily ensured by the different inputs obtained through the utilization of the k-means method. This approach achieves data diversity. Other studies, such as \cite{jiang2021distributed,polyzos2021policy}, have made advancements towards enhancing the composition of RL.} Here, most studies use the offline algorithm, except for Polyzos et al. \cite{polyzos2021policy}, who used an online algorithm. The performance of the ERL method has been verified on simulation platforms \cite{sadeghi2017optimal}. It can be seen from experimental results that the use of ensemble models makes the ERL method schedule significantly better than compared RL methods. When applying ERL in this area, matching the application requirements of multi-agent and distributed architecture becomes a core point. This system architecture allows the ensemble models in ERL to handle the same or different tasks.

\begin{table}[htp]
\centering
\caption{Application in IoT and cloud computing area}
\label{Application in IoT and cloud computing area}
\begin{tabularx}{\textwidth}{p{1cm}p{4.5cm}XX}
\toprule[1.5pt]
Year&Authors& Problem& Training Algorithm \\
\midrule[1pt]
2020 & Ashiquzzaman et al. 
\cite{ashiquzzaman2020energy} & IoT sensor calibration  & deep Q-Network     \\
2021 & Polyzos et al. 
\cite{polyzos2021policy}          
& resource allocation    
& Sarsa              \\
2021 & Jiang et al.  \cite{jiang2021distributed}          & large-scale MEC systems & deep Q-Network     \\
2021 & Gu et al.  
\cite{gu2021heterogeneous}   
& online cloud task scheduler                                 
& deep deterministic policy gradient \\
2021 & Liu et al.   \cite{liu2021elegantrl}      & deep reinforcement learning   training on GPU cloud platform & actor-critic network               \\
2022 & Mahmud et al.  
\cite{mahmud2022ensemble} & non orthogonal multiple access   unmanned aerial network     & deep Q-Network\\
\bottomrule[1.5pt]
\end{tabularx}
\end{table}

\textcolor[rgb]{0,0,0}{For the domains of IoT and cloud computing, the evaluation metrics exhibit some variations. The performance of the ERL method for IoT sensor calibration in \cite{ashiquzzaman2020energy} is based on accuracy. Unlike \cite{ashiquzzaman2020energy}, \cite{polyzos2021policy} examines the proposed method's effectiveness by analyzing the total cost of IoT resource allocation. In terms of the cloud computing domain, workload serves as a crucial evaluation metric \cite{gu2021heterogeneous}.}

\subsection{Financial Area}

In the financial area, complex decision-making problems, such as pricing financial products and portfolio optimization,  are being tried to be solved by ERL methods. Though single models can make predictions on a specific problem, their generalization is affected by the problem scenario. Compared with the single model, ensemble models are affected less by the problem scenario factors. Table \ref{Application in financial area} presents the applications of ERL methods in finance. In these studies, 67\% used only one training algorithm, while the rest of the studies used multiple training algorithms in an ERL method. Three algorithms, namely proximal policy optimization, advantage actor-critic, and deep deterministic policy gradient, have shown good performance in training.

\begin{table}[htp]
\centering
\caption{Application in financial area}
\label{Application in financial area}
\begin{tabularx}{\textwidth}{p{1cm}p{4cm}XX}
\toprule[1.5pt]
Year & Authors                                                    
& Problem                 
& Training Algorithm \\
\midrule[1pt]
2020 &
Yang et al.  \cite{yang2020deep} &
stock trading &
proximal policy optimization,   advantage actor-critic, deep deterministic policy gradient \\
2020 & Xu et al. \cite{xu2020ensemble}             & fuel economy improvement & Q-learning         \\
2021 & Carta et al.   
\cite{carta2021multi}       & stock market forecasting & deep Q-Network     \\
2022 & Li et al.  \cite{li2022new}                 & regional GDP prediction  & deep Q-Network     \\
2022 & Zijie Cao and Hui Liu  
\cite{cao2022novel} & carbon price forecasting & Q-learning       
\\
2022 &
N{\'e}meth, Marcell and Sz{\H{u}}cs, G{\'a}bor \cite{nemeth2022split} &
algorithmic trading &
proximal policy optimization,   advantage actor-critic, deep deterministic policy gradient 
\\
\bottomrule[1.5pt]
\end{tabularx}
\end{table}

\textcolor[rgb]{0,0,0}{The utilization of multiple training algorithms constitutes a fundamental strategy employed by ERL to uphold diversity in financial domains \cite{yang2020deep,nemeth2022split}. Specifically, the base learner is trained to obtain different parameter configurations for parameter diversity. The primary objective of ERL is prediction, whereby cumulative return, annualized return, annualized volatility, Sharpe ratio, and max drawdown emerge as five commonly adopted metrics for stock trading evaluation \cite{yang2020deep}. As a forecasting task within the domain of energy environment studies, MAE, MAPE, and SDE can also be employed to evaluate method performance \cite{li2022new}. Additionally, Theil U statistic 1 (U1) can be utilized as an evaluative metric \cite{cao2022novel}.}

\subsection{Other Areas }

Apart from the previous three traditional application domains, ERL has also been successfully applied in various other fields such as transportation, medicine, and security, which will be discussed in detail within this section. Table \ref{Application in other areas} provides a comprehensive overview of these ERL methods primarily focus on prediction tasks while only a limited number of classification problems like diagnosis and recognition are addressed using ensemble techniques. Notably, the work conducted by Eriksson et al. \cite{eriksson2022sentinel}, who employed ERL methods to tackle autonomous driving challenges, deserves special attention. If the ERL can be effectively implemented for small-scale autonomous driving assistance systems, it is highly likely to stimulate new research endeavors and practical applications in this domain.

\begin{table}[htp]
\centering
\caption{Application in other areas}
\label{Application in other areas}
\begin{tabularx}{\textwidth}{p{1cm}p{4cm}XX}
\toprule[1.5pt]
Year & Authors                                                     
& Problem                      & Area     \\
\midrule[1pt]
2021 & Ghosh et al. 
\cite{ghosh2021deep}          
& air traffic control         
& traffic  \\
2021 & Dong et al. 
\cite{dong2021novel}           
& traffic speed forecasting   
& traffic  \\
2022 & Shang et al. 
\cite{shang2022new}           
& traffic volume forecasting  
& traffic  \\
2022 & Qi et al.  \cite{qi2022random}               & traffic signal control       & traffic  \\
2022 & Eriksson et al. 
\cite{eriksson2022sentinel} & autonomous driving           & traffic  \\
2016 & Tang et al. 
\cite{tang2016inquire}         
& symptom checker             
& medicine \\
2021 & Jalali et al. 
\cite{jalali2021oppositional} & COVID-19 diagnosis           & medicine \\
2022 & Birman et al. 
\cite{birman2022cost}        
& malware detection           
& security \\
2022 & Li et al.    \cite{li2022deep}               & rumor tracking               & security \\
2023 & Henna et al. 
\cite{henna2023ensemble}      
& FSO/RF communication systems & optics   \\
2019 & Cuay{\'a}huitl et al. 
\cite{cuayahuitl2019ensemble} & chatbots                           
& dialogue system     \\
2010 & Alexander Hans and Steffen Udluft  
\cite{hans2010ensembles}         
& pole balancing                     
& engineering control \\
2018 & Ferreira et al.  
\cite{ferreira2018multiobjective}                   & cognitive satellite   communication & aerospace        \\
\bottomrule[1.5pt]
\end{tabularx}
\end{table}

In the future, we expect to see more areas using ERL methods for complex tasks. Existing research results can provide valuable references for subsequent research, including improving existing algorithms to overcome their limitations, or extending the problem domain to obtain new insights. In the next section, we discuss some potential directions for future research on ERL methods.

\section{Datasets and Compared Methods}
\label{dataset}
This section examines the datasets and comparison methods used in various studies related to ensemble reinforcement learning (ERL). As presented in Table \ref{Datasets and compared Methods}, experiments are conducted to evaluate the performance of the proposed ERL methods. The datasets used in these experiments mainly include real-world data and publicly available datasets or environments. Real-world data are useful for objectively testing the predictive or classification performance of the method for specific applications. For instance, studies in the field of energy and environment have gathered data from multiple cities to predict desired outcomes \cite{chengqing2023multi, liu2020new}. In contrast, publicly available datasets or environments such as the OpenAI Gym environment in the field of reinforcement learning are widely used to test the predictive performance of algorithms for continuous/discrete actions \cite{brockman2016openai}. The UCI machine learning repository is widely recognized as the predominant public dataset for classification problems in academic research \cite{liu2020instance}. Furthermore, some medical-specific datasets are also utilized in studies of disease diagnosis \cite{jalali2021oppositional}.

\begin{table}[htp]
\tiny
\centering
\caption{Datasets and compared Methods}
\label{Datasets and compared Methods}
\begin{tabularx}{\textwidth}{p{1cm}p{2cm}Xp{8cm}}
\toprule[1.5pt]
Year & Authors & Dataset & Compared Methods \\
\midrule[1pt]
2016 &
Osband et al.   \cite{osband2016deep} &
Atari games &
DQN \\
2017 &
Chen et al. \cite{chen2017ucb} &
Atari games &
A3C+ \\
2017 &
Partalas et al.   \cite{partalas2009pruning} &
UCI machine learning repository &
classifier combination methods voting (V) and SMT and the forward selection (FS), selective fusion (SF) \\
2018 &
Pearce et al.   \cite{pearce2018bayesian} &
Cart Pole control problem &
Q-learning with different layer   NNs \\
2019 &
Dong et al. \cite{dong2021novel} &
traffic speed dataset &
GRU, LSTM, MLP, RBF, LSTM-GRU-GA \\
2019 &
Pan et al. \cite{pan2019ensemble} &
Maze, Mountain Car, Robotic   Soccer Game Simulation &
counterpart \\
2019 &
Goyal et al.   \cite{goyal2019reinforcement} &
CATS (Competition on Artificial   Time series) dataset &
LSTM, ANN, Linear regression,   Random Forest, Online NN \\
2019 &
Macheng Shen and Jonathan P How   \cite{shen2021robust} &
two-player asymmetric game &
single model, RNN \\
2020 &
Qingfeng Lan et al.   \cite{lan2020maxmin} &
Mountain Car &
Q-learning, Double Q-learning,   Averaged Q-learning \\
2020 &
Liu et al. \cite{liu2020new} &
three different groups of   measured wind speed data from Xinjiang wind farms &
Network: LSTM method, the DBN method, the ESN method; Training algorithm: SARSA \\
2020 &
Lin et al. \cite{lin2020ensemble} &
Maze, soccer robot game &
orthogonal projection inverse reinforcement learning method (OP-IRL) \\
2020 &
Junta Wu and Huiyun Li   \cite{wu2020deep} &
2D Robot Arm  Open Racing Car Simulator (TORCS) &
DDPG \\
2020 &
Yang et al. \cite{yang2020deep} &
Dow Jones 30 constituent stocks   (at 01/01/2016) &
PPO, A2C, DDPG \\
2020 &
Liu et al. \cite{liu2020instance} &
UCI online data repository &
classifiers combination approaches majority voting (MV), weighted voting (WV), ensemble selection methods forward selection (FS) \\
2021 &
Ghosh et al. \cite{ghosh2021deep} &
open-source air trafﬁc simulator &
PPO \\
2021 &
Jalali et al. \cite{jalali2021new} &
GHI data sets &
adaptive hybrid model (AHM),   hybrid feature selection method (HFS), Outlier-robust hybrid model (ORHM),   novel hybrid deep neural network model (NHDNNM), OHS-LSTM \\
2021 &
Liu et al. \cite{liu2021new} & data collected from a congested  intersection in Changsha &
RNN, ENN, ESN, DBN, RBF, GRNN, MLP \\
2021 &
Jalali et al.   \cite{jalali2021oppositional} &
two well-known open-source image datasets named Mendely and Kaggle &
the original version of GSK and eight powerful evolutionary algorithms including grasshopper optimization algorithm (GOA),   Slime mold algorithm (SMA), genetic algorithm, gray wolf optimizer (GWO),   particle swarm optimization (PSO), differential evolution (DE), biogeography-based optimization (BBO) \\
2022 &
Hassam Ullah Sheikh et al.   \cite{sheikh2022maximizing} &
Mujoco environments, Atari games &
TD3, SAC and REDQ \\
2022 &
Shang et al. \cite{shang2022new} &
actual traffic volume data of nine stations of Changsha freeway & Chebnet, CNN, LSTM, DBN, RNN, ESN, multi-layer perceptron (MLP) \\
2022 &
Tan et al. \cite{tan2022new} &
actual data &
RNN, the deep belief network   (DBN), the echo state network (ESN), the error encoding network (ENN),   General Regression Neural Network (GRNN), radial basis function network   (RBF), multilayer perceptron (MLP) \\
2022 &
Li et al. \cite{li2022new} &
three sets of data from three   Provinces of China &
ESN, ENN, RNN, BPNN, ELM, RBF \\
2022 &
Cao et al. \cite{cao2022novel} &
The data for the three carbon trading markets come from the Hubei Carbon Trading Network, Beijing Carbon Emissions Electronic Trading Platform, and International Carbon Action Partnership (ICAP) & Network: TCN, BiLSTM, KELM, BPNN, MLP, echo state network (ESN), Elman neural network (ENN), and gradient boosting   decision tree (GBDT); Training algorithm: SARSA \\
2022 &
Qin et al.   \cite{qin2022optimization} &
historical load data of the California Independent System Operator (CASIO)  from January 1, 2021 to July 5, 2021 &
PPO guided tree search, the MIQP algorithm   with Gurobi 9.1 \\
2022 &
Sogabe et al.   \cite{sogabe2022attention} &
optimal energy management in a   residential building microgrid &
mixed-integer linear programming   (MILP) \\
2022 &
Birman et al.   \cite{birman2022cost} &
a range of real-world scenarios &
Aggregation method \\
2022 &
Li et al. \cite{li2022deep} &
PHEME, RumorEval &
Naive Bayes, SVM-SGD, Dense,   BiLSTM, FastText, TextCNN, VRoC, some combinations of the above methods \\
2022 &
Sharma et al.   \cite{sharma2022deepevap} &
two MEC servers and 30 IoTDs randomly   distributed in the squared area with size 50m×50m &
Actor-Critic, DDPG \\
2022 &
Schubert et al.   \cite{schubert2022polter} &
SymCat’s symptom-disease   database &
single model-based RL \\
2023 &
Yu et al.   \cite{chengqing2023multi} &
actual wind power data of nine wind turbines &
GMDH, DBN, ESN, ENN, the extreme   learning machine (ELM), the radial basis function (RBF), multi-layer perceptron\\
\bottomrule[1.5pt]
\end{tabularx}
\end{table}

To assess the efficacy of the proposed ERL methods, various comparative approaches have been employed in existing literature. Among these, the single model-based RL method (SM-RL) is one of the simplest ways to reflect the effectiveness of the proposed ERL method \cite{schubert2022polter}. The training algorithm used in SM-RL remains consistent with that of the ERL method. However, this compared method has limited convincing power. Consequently, some other studies have used other training algorithms to compare with the proposed algorithm from another perspective \cite{sheikh2022maximizing, ghosh2021deep}. In order to comprehensively evaluate the effectiveness of algorithms, it is necessary to separately assess different models, training algorithms, and integration methods \cite{liu2020new}.

\textcolor[rgb]{0,0,0}{The comparison methods are continuously evolving through ongoing research. In other words, existing ERLs proposed in the relevant literature serve as baselines or state-of-the-art (SOTAs) for comparison when introducing a new ERL method. However, reproducing the exact method described in the literature may pose challenges due to various influencing factors such as the environment and algorithm parameters. To address this issue, many journals or conferences now require disclosure of datasets, pseudo-code, code, model structure, hyper-parameter configuration, data partitions, tuning methods and statistical tests as basic requirements. Studies like \cite{partalas2006ensemble,pak2020deep,wang2021federated} have provided detailed information on their models and training methods which greatly assist other researchers who consider them as SOTA methods. Nevertheless, some studies still lack certain details (e.g., random seeds and training strategies), resulting in reproduced results that fall short of expectations. This discrepancy arises from the fact that ERL is a class of improved RL methods where models can vary depending on the environment.}

\section{Open Questions and Future Research Directions}
\label{open question}
\subsection{Open Questions}
\begin{figure}[htp]
\centering
\includegraphics[width=0.5\textwidth]{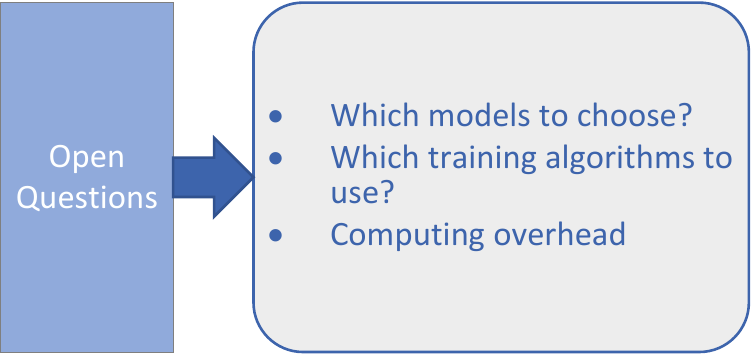}
\caption{\textcolor[rgb]{0,0,0}{Open Questions. These three open questions are important and worthy of in-depth study for the development of ERL.}}
\label{open questions}
\end{figure}

In this section, we summarize three open questions in ERL-based research that can contribute to the future development of ERL (see Figure \ref{open questions}).

\subsubsection{Which Models to Choose?}

Models serve as the foundation for constructing ERL methods and exert a direct influence on the ultimate outcome. The utmost crucial aspect of model selection lies in its capability to perform feature selection and learning effectively. If a model cannot acquire valid information, it becomes devoid of meaning. Compared to single-model methods, ensemble methods employing multiple identical or diverse models can mitigate the risk of erroneous inference and enhance overall predictive performance. Consequently, it is imperative that the implemented models contribute significantly to the overall predictive performance of the ERL method.

The models implemented in ERL mainly include ML and ANN models, characterized by their simplistic structure and robust generalization capabilities, rendering them suitable for various practical applications. However, when confronted with large-scale data and numerous features, ML models may encounter challenges. In such scenarios, the ANN model excels at extracting features from datasets and generating predictions that closely align with actual values. Nevertheless, it is worth noting that certain tasks requiring comprehension pose difficulties for ANN models. To address this limitation, some studies such as \cite{saadallah2021online}, and \cite{birman2022cost} have explored the integration of both ML and ANN models within ERL methods.

Simultaneously utilizing models at different training stages greatly facilitates the design of ERL \cite{carta2021multi}, as these models exhibit variations in parameters and distinct predicted bias and variance. When employing such an ensemble model, it is worthwhile to thoroughly investigate the conditions under which each ensemble element is preserved during the training process.

To automate the optimization of model structures, some studies have employed proposed ensemble pruning methods such as forward selection (FS) \cite{caruana2004ensemble} and selective fusion (SF) \cite{tsoumakas2005selective}. These studies utilize a model library for selecting models, thereby reducing human effort in the selection process. The model library can also be continuously extended based on the latest research to optimize the overall performance of the ERL.

\subsubsection{Which Training Algorithms to Use?}

Model training poses a well-known challenge in the ERL method. For model-free based ERL, the agent updates model parameters based on state transfer or trajectory after a certain number of iterations. However, there is no guarantee that all sampled data are useful. The classic strategy is to use experience delay technology, which can improve the sampling efficiency. While Schaul et al. \cite{ schaul2015prioritized} found that the sampling inefficiency problem occurs when the sample in the relay buffer is useless. Selected experience relay makes the algorithm training more efficient by selectively choosing the adopted data into the replay buffer \cite{isele2018selective}. It is worth noting that the experience relay buffer is only applicable to the off-policy RL method. Besides, model-based RL methods can guarantee sampling efficiency by learning environment models. However, such a training algorithm has to face a huge action space. In this case, approximating the environment model becomes an extremely difficult task.

Good training algorithms should balance exploration and exploitation, as relying solely on one approach proves challenging. Therefore, the future direction of development lies in employing multiple types of training algorithms simultaneously. This ERL method necessitates designing separate sampling techniques for the solution space based on the training strategy. Furthermore, it is crucial to consider how to use sampled data from the same strategy for model training processes. In addition, such algorithmic training imposes significant demands on the computational capabilities of both CPU and GPU.

\subsubsection{Computing Overhead}

Computing overhead is a closely related issue that must be taken into consideration for ERL, in addition to the aforementioned problems. Implementing multiple ML or ANN models in ERL results in an ultra-large number of parameters compared to a single model. Particularly when each ANN model possesses a complex structure, memory consumption becomes an indispensable factor. Similarly, multiple training algorithms can complicate the training process. Even with computational acceleration techniques, the time required for numerous computations can significantly exceed that of individual training sessions. Many studies have found that ERL methods can complete sampling efficiently, but are also accompanied by an increase in computation time \cite{sheikh2022maximizing,polyzos2021policy}. In the testing stage, complex decisions then easily lead to longer computation time than other methods \cite{ partalas2009pruning}. So, some researchers tried to design strategies based on scenarios, which reduce the computational overhead to some extent. An et al. achieved a reduction in model training time along with a reduction in memory consumption by taking uncertainty into account in the ERL method \cite{ an2021uncertainty}. Pan et al. reduced the time consumed by the algorithm for each iteration of training by fuzzifying the reward \cite{pan2019ensemble}. Up to now, the number of computationally cost-reducing models is still small. So, it is difficult to show that the improvement strategy is still applicable to large-scale models. In addition, the cost of data interaction needs extra attention when the models are deployed on multiple machines, which will affect the efficiency of the system.

The cost-effectiveness of ERL using complex structures and training processes is a fundamental basis for measuring method design and algorithm training. Increasing the number of models can improve the ensemble prediction performance but is also accompanied by an increase in computing overhead. After a certain number of models are implemented, the computing overhead of using more models can be significantly greater than the improvement in method performance. In such cases, increasing the size of the ensemble model is not advisable.

In certain practical application problems, the feasible solutions obtained through ERL methods can as problem-solving outcomes. Controlling the number of iterations of the training process is an alternative if the computing overhead of searching for the optimal strategy is much greater than the contribution it can make. Thus, addressing the issue of computing overhead is essential for the successful application of ERL in various fields.

\subsection{Future Research Directions}
ERL has been extensively used in valuable research to effectively address scientific problems and various application domains. Based on the analysis of relevant literature mentioned in the survey, it is evident that a majority of the research has been concentrated within the past decade. Numerous unexplored research directions await investigation by scholars, and this section presents several potential directions for further inquiry. 

1. \textbf{Randomized models}: Randomized models, such as random vector functional link networks \cite{pao1994learning}, random initialized implicit layers, \cite{shi2021random} have emerged as effective strategies for training reduction. In addition, the utilization of implicit/explicit ensembles \cite{han2017branchout} can improve the model training efficiency from the perspective of diversity. Ensuring diversity among base learners is a core problem that needs to be solved in the ERL method and deserves further study.

2. \textbf{Effect of decision strategy}: Decision strategies are employed to derive the final output based on the predictive outcomes of individual base learners. Despite various attempts made in previous studies to use different types of decision strategies, there remains a dearth of systematic research investigating their impact. Accordingly, it is imperative to conduct a comprehensive analysis of the decision strategies applicable in various integrated models and the number of training algorithms.

3. \textbf{Hierarchical ensemble}: Hierarchical reinforcement learning methods have been used to solve some challenging problems. For example, Qin et al. and Ferreira et al., respectively, endeavored to employ multiple RL models to complete different tasks separately in order \cite{tan2022new,qin2022hrl2e,ferreira2018multiobjective}. The current model structure is designed based on empirical knowledge and lacks systematic theoretical validation, which calls for further investigation. In the context of hierarchical ensemble reinforcement learning, it is crucial to carefully evaluate the performance of both individual RL models and ensemble models. Additionally, a meticulous design should be employed to determine the specific role of each element within the hierarchical framework in order to address specific problems.

4. \textbf{Large-scale ensemble}: Existing ERL methods typically employ around three base learners \cite{perepu2020reinforcement,elliott2021wisdom}. From the diversity perspective, incorporating a larger number of models in a new ERL method can effectively explore extensive feature information and enhance prediction accuracy. From a statistical perspective, increasing the number of ensemble components enables the generation of more hypotheses and enhances the likelihood of identifying the optimal hypothesis. By employing a large-scale ERL, it is possible to design an information-sharing mechanism, thereby reducing the total training cost of the model. 

5. \textbf{Distributed approach}: Ensemble reinforcement learning can also be trained or used in a distributed manner. Existing research on distributed ERL primarily focuses on its implementation within a distributed framework and lacks methodological advancements  \cite{jiang2021distributed,eriksson2022sentinel}. Therefore, further analysis is warranted on how to effectively leverage both ERL and distributed reinforcement learning. Integrating ERL into a distributed framework inevitably incurs augmented costs associated with model training and communication. Hence, in a distributed framework, it becomes imperative to prioritize low-cost training methods and controlled training time to ensure the practical applicability of ERL. 

6. \textbf{Online model training}: Currently, ERL adopts offline training and direct online implementation. However, this model training algorithm poses challenges in capturing the most up-to-date information, thereby affecting the optimality of the agent's strategy. To address this limitation, incorporating online or near-online model training methods would enable the timely addition of new information to the training dataset, ensuring that the model can effectively respond to dynamic situations. It is crucial for online training to focus on establishing a triggering mechanism for model updates as both over-training and under-training can adversely impact the performance of the ERL method. Moreover, forgetting history memory can facilitate ERL in discovering novel optimal strategies.

7. \textbf{Efficient training}: The sampling efficiency of DRL also deserves attention, as it remains a prevalent issue in ERL methods. Consequently, this gives rise to several associated challenges, including data set partitioning for training, model parameter initialization, hyperparameter tuning, and strategy updating. Models from different training stages can also be combined to identify the optimal configuration of model settings \cite{ carta2021multi}.

8. \textbf{Embedded into big data platform}: Although most of the current studies on ERL are conducted in simulation environments, there still exists a gap between these findings and their practical application. To address this issue, integrating ERL methods into a big data platform enables timely inference based on system-acquired data for various practical forecasting tasks. For both short-term forecasting and long-term forecasting objectives, diverse ERL methods can be deployed within the big data platform.

\textcolor[rgb]{0,0,0}{9. \textbf{Hyperparametric reduction}:  The integration of multiple training algorithms in ERL can lead to a significant proliferation of hyperparameters within the method, thereby rendering model training arduous. Moreover, the trained model may exhibit instability when the scenario changes \cite{zhang2021importance}. Hence, it is imperative to propose a mechanism for reducing hyperparameter complexity and alleviating the burden on researchers involved in hyperparameter tuning. One solution idea to solve this problem is to refer to the idea of pre-training by completing a rapid deployment of the model first. Subsequently, the deployed model is simply tuned to improve performance.}

The above aspects provide potential directions for future research, although they are not exhaustive. It is important to acknowledge that the design and problem-solving processes of ERL methods may encounter various new situations. Furthermore, it should be noted that the no free lunch theorem applies universally to all ERL methods \cite{sutton2018reinforcement}. Therefore, when designing ERL methods, careful consideration must be given to their complexity and training time requirements.

\section{Conclusion}
\label{conclusion}
This paper presents a comprehensive review of the research progress on ERL methods, covering various aspects including background, strategies, applications, and future directions. Firstly, the description of RL methods and EL methods has enhanced the understanding of ERL. Secondly, various strategies, such as Q-function ensemble, model combination, and decision strategies, are introduced and discussed. Subsequently, the application of ERL methods, datasets, and compared methods are described. Additionally, future research directions that can further enhance the performance of ERL have been extensively discussed.

The robust predictive and classification capabilities of ERL render it a promising framework for addressing intricate problems. E demonstrated successful applications across diverse domains, encompassing finance, robotics, and healthcare. Nevertheless, there exists substantial potential for future research endeavors. This paper highlights potential research directions including randomized models, the impact of decision strategies, hierarchical ensembles, large-scale ensembles, distributed approaches, online model training techniques, efficient training methodologies, and integration of ERL into big data platforms.

The future holds promising prospects for ERL performance across a wider range of application domains. Consequently, it is imperative for researchers to persistently explore and develop novel ERL methods that effectively tackle the challenges encountered in practical scenarios.

\paragraph{Acknowledgement}
This work is supported by the National Natural Science Foundation of China (72201273,72001212), the Science and Technology Innovation Team of Shanxi Province (2023-CX-TD-07), the Special Project in Major Fields of Guangdong Universities (2021ZDZX1019), and the Hunan Key Laboratory of Intelligent Decision-making Technology for Emergency Management (2020TP1013).

\bibliography{mybib}{}
\bibliographystyle{elsarticle-num}

\end{document}